\documentclass[letterpaper, 10 pt, conference]{packages/ieeeconf}  

\IEEEoverridecommandlockouts                              

\usepackage{times}
\usepackage{epsfig}
\usepackage{graphicx}
\usepackage{amsmath}
\usepackage{amssymb}
\usepackage{subcaption} 
\usepackage[ruled]{algorithm}
\usepackage{algpseudocode}
\usepackage{ctable}
\usepackage[export]{adjustbox}
\usepackage{import}
\usepackage{pgf}
\usepackage{setspace}
\usepackage{url}
\usepackage{hyperref}
\usepackage{capt-of,etoolbox}
\usepackage{pifont} 
\newcommand{\xmark}{\ding{55}} 

\makeatletter
\let\NAT@parse\undefined
\makeatother
\usepackage[numbers]{natbib}

\usepackage{packages/sparo_acronyms}
\usepackage{packages/sparo_math}
\usepackage{packages/sparo_SIunits}
\usepackage{packages/sparo_misc}
\usepackage{multirow}
\usepackage{multicol}
\usepackage{indentfirst}
\usepackage{float}
\usepackage{soul,color}
\usepackage{verbatim} 

\usepackage{xcolor}

\newcommand{\rl}[1]{{\textcolor{red}{#1}}}

\definecolor{gl}{HTML}{008000}
\newcommand{\gl}[1]{{\textcolor{gl}{#1}}}

\usepackage[symbol]{footmisc}

\let\oldtwocolumn\twocolumn
\renewcommand\twocolumn[1][]{%
    \oldtwocolumn[{#1}{
    \begin{center}
           \includegraphics[clip, trim= 0 125 0 130 ,width=0.90\textwidth]{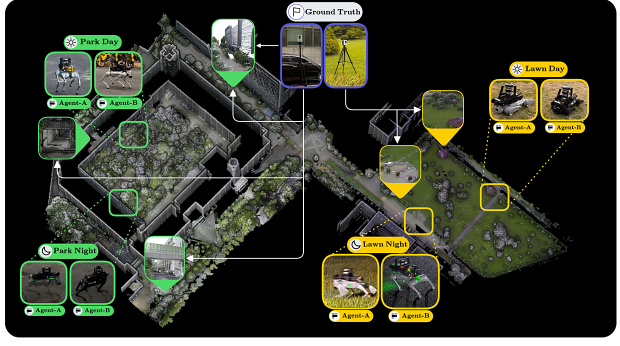}
           \vspace{-0.2cm}
           \captionof{figure}{Survey-grade prior map from DiTer++ dataset containing various scenes (e.g. buildings, curbstones, and uneven terrain).
                              Multiple legged robots equipped with a multi-modal sensor configuration traverse the prior map in multi-session (i.e., day and night) regardless of terrain.}
           \label{fig:main}
        \end{center}
    }]
}

\begin{document}

\title{\LARGE \bf
   DiTer++: Diverse Terrain and Multi-modal Dataset \\ for Multi-Robot SLAM in Multi-session Environments
}

\author{Juwon Kim$^{1}$, Hogyun Kim$^{1}$, Seokhwan Jeong$^{1}$, Youngsik Shin$^{2}$, and Younggun Cho$^{1\dagger}$
	\thanks{$^{1}$Juwon Kim, $^{1}$Hogyun Kim, $^{1}$Seokhwan Jeong, and $^{1\dagger}$Younggun Cho are with the Electrical and Computer Engineering, Inha University, Incheon, South Korea 
		{\tt\small marimo117@inha.edu, [hg.kim, eric5709]@inha.edu, yg.cho@inha.ac.kr} \newline $^{2}$Youngsik Shin$^{2}$ is with Depart. of AI Machinery, Korea Institute of Machinery and Materials, Daejeon, South Korea {\tt\small yshin86@kimm.re.kr}}%
}

\maketitle

\begin{abstract} 
We encounter large-scale environments where both structured and unstructured spaces coexist, such as on campuses.
In this environment, lighting conditions and dynamic objects change constantly.
To tackle the challenges of large-scale mapping under such conditions, we introduce DiTer++, a diverse terrain and multi-modal dataset designed for multi-robot SLAM in multi-session environments.
According to our datasets' scenarios, \texttt{Agent-A} and \texttt{Agent-B} scan the area designated for efficient large-scale mapping day and night, respectively.
Also, we utilize legged robots for terrain-agnostic traversing. 
To generate the ground-truth of each robot, we first build the survey-grade prior map.
Then, we remove the dynamic objects and outliers from the prior map and extract the trajectory through scan-to-map matching.
Our dataset and supplement materials are available at \url{https://sites.google.com/view/diter-plusplus/}. 
\end{abstract}

\section{Introduction}
To increase the efficiency and scalability of mapping, multi-robot \ac{SLAM} plays a groundbreaking role.
However, we often encounter diverse terrain environments where hazards and dynamic objects are scattered.
Also, these environments with tall buildings or dense trees hinder a robots' precise localization.

\begin{table*}[t]
\caption{Various Multi-Robot and Diverse Terrain Datasets}
\renewcommand{\arraystretch}{1.2}
\centering\resizebox{\textwidth}{!}{
{
\begin{tabular}{c||cccccccccc}
\toprule[1.5pt]
\textbf{Dataset}                          & \textbf{Ground Truth} & \textbf{Multi-robot} & \textbf{Multi-session} & \textbf{LiDAR}    & \textbf{RGB}
                                          & \textbf{Thermal}      & \textbf{IMU}         & \textbf{Robot Joint}   & \textbf{Hardware} & \textbf{Environments}  \\ \hline
DiTer \cite{jeong2024diter}               & \textbf{G}            & \rl{\xmark}          & \rl{\xmark}            & \gl{\checkmark}   & \gl{\checkmark} 
                                          & \gl{\checkmark}       & \gl{\checkmark}      & \gl{\checkmark}        & Legged            & Diverse Terrain         \\ 
KITTI \cite{geiger2012we}                 & \textbf{G}            & \rl{\xmark}          & \rl{\xmark}            & \gl{\checkmark}   & \gl{\checkmark}
                                          & \rl{\xmark}           & \gl{\checkmark}      & \rl{\xmark}            & Vehicle           & Urban                   \\
STheReO \cite{yun2022sthereo}             & \textbf{G/L}          & \rl{\xmark}          & \gl{\checkmark}        & \gl{\checkmark}   & \gl{\checkmark}   
                                          & \gl{\checkmark}       & \gl{\checkmark}      & \rl{\xmark}            & Vehicle           & Urban                   \\
ViViD++ \cite{lee2022vivid++}             & \textbf{G/L/M}        & \rl{\xmark}          & \gl{\checkmark}        & \gl{\checkmark}   & \gl{\checkmark}   
                                          & \gl{\checkmark}       & \gl{\checkmark}      & \rl{\xmark}            & Vehicle           & Indoor/Urban            \\
TAIL \cite{wang2024we}                    & \textbf{G}            & \rl{\xmark}          & \rl{\xmark}            & \gl{\checkmark}   & \gl{\checkmark}    
                                          & \rl{\xmark}           & \gl{\checkmark}      & \gl{\checkmark}        & Legged            & Diverse Terrain          \\
DISCO SLAM \cite{huang2021disco}          & \textbf{G}            & \gl{\checkmark}      & \rl{\xmark}            & \gl{\checkmark}   & \rl{\xmark}   
                                          & \rl{\xmark}           & \gl{\checkmark}      & \rl{\xmark}            & Wheeled           & PARK                     \\
S3E \cite{feng2022s3e}                    & \textbf{G}            & \gl{\checkmark}      & \rl{\xmark}            & \gl{\checkmark}   & \gl{\checkmark}   
                                          & \rl{\xmark}           & \gl{\checkmark}      & \rl{\xmark}            & Wheeled           & Campus                   \\
Kimera Multi \cite{tian2023resilient}     & \textbf{G/L}          & \gl{\checkmark}      & \rl{\xmark}            & \gl{\checkmark}   & \gl{\checkmark}   
                                          & \rl{\xmark}           & \gl{\checkmark}      & \rl{\xmark}            & Wheeled           & Campus                    \\
Hetero \cite{chen2024heterogeneous}       & \textbf{G/S}          & \gl{\checkmark}      & \gl{\checkmark}        & \gl{\checkmark}   & \gl{\checkmark}
                                          & \rl{\xmark}           & \gl{\checkmark}      & \rl{\xmark}            & Wheeled           & Tunnel/Water Surface      \\
Nebula Multi \cite{chang2022lamp}         & \textbf{S}            & \gl{\checkmark}      & \rl{\xmark}            & \gl{\checkmark}   & \rl{\xmark}    
                                          & \rl{\xmark}           & \rl{\xmark}          & \rl{\xmark}            & Legged            & Underground               \\
BotanicGarden \cite{liu2024botanicgarden} & \textbf{S}            & \rl{\xmark}          & \rl{\xmark}            & \gl{\checkmark}   & \gl{\checkmark}    
                                          & \rl{\xmark}           & \gl{\checkmark}      & \gl{\checkmark}        & Wheeled           & PARK                        \\
MCD \cite{nguyen2024mcd}                  & \textbf{S}            & \rl{\xmark}          & \gl{\checkmark}        & \gl{\checkmark}   & \gl{\checkmark}    
                                          & \rl{\xmark}           & \gl{\checkmark}      & \rl{\xmark}            & Wheeled/Handheld  & Campus                  \\ \hline
\textbf{Ours (DiTer++)}                   & \textbf{S}            & \gl{\checkmark}      & \gl{\checkmark}        & \gl{\checkmark}   & \gl{\checkmark} 
                                          & \gl{\checkmark}       & \gl{\checkmark}      & \gl{\checkmark}        & Legged            & Diverse Terrain         \\ \bottomrule[1.5pt]
\multicolumn{11}{r}{\textbf{S}, \textbf{G}, \textbf{L}, \textbf{M} represent the survey-grade prior map, GPS, LiDAR SLAM, and motion capture system, respectively.} \\
\end{tabular}}}
\vspace{-0.6cm}
\label{tab:related}
\end{table*}

In this paper, we propose a new multi-robot, multi-session, and multi-modal dataset utilizing legged-robots that evolved from our previous work \cite{jeong2024diter}. 
Instead of various multi-robot datasets \cite{huang2021disco, feng2022s3e, tian2023resilient, chen2024heterogeneous, chang2022lamp} focused on a single session and task, we provide a multi-robot dataset of different sensors acquired at a different session.
The main contributions of this work are the following:
\begin{itemize}
   \item \textbf{Survey Level Ground Truth:} 
        To address the attenuation and ambiguity of \ac{GPS} signals encountered in previous work \cite{jeong2024diter}, we first generate a survey-grade prior map and extract pose ground truth through scan-to-map matching. Additionally, we remove dynamic objects and detect changes within the prior map to provide a more accurate ground-truth trajectory, as shown in \figref{fig:main}.
    \item \textbf{Multi-robot:}   
        We utilize multiple legged-robots to traverse diverse terrains such as lawns, curbs, and indoor environments. 
        In our dataset's scenario, each robot performs efficient mapping by dividing the mapping area with partial overlap.
    \item \textbf{Multi-session:} 
        The environment presents various challenges, such as fluctuating lighting conditions caused by buildings and street lamps. To address this, we provide multi-session data, distinguishing between day and night sessions.
    \item \textbf{Multi-modal:}   
        Our dataset is collected using multiple perceptual sensors, including an RGB camera, \ac{LiDAR}, and a \ac{TIR} camera for capturing information beyond the visible spectrum. In addition, we provide motion data from built-in joint sensors and \ac{IMU}, which assist in navigation and \ac{SLAM} tasks.
\end{itemize}

\section{Related Works}
Our dataset is acquired via legged-robots in diverse terrain environments including three scenarios (i) multi-robot, (ii) multi-session, and (iii) multi-modal.
Additionally, each sequence provides reliable ground-truth data.
In this section, we review and analyze various influential datasets from the perspective of these scenarios, including their methods for generating ground-truth, as summarized in \tabref{tab:related}.

\subsection{Datasets for Multi-Robot, Session, and Modal}
\begin{itemize}
    \item \textbf{Multi-robot} datasets were proposed to facilitate the testing of algorithms such as multi-robot \ac{SLAM} \cite{huang2021disco, feng2022s3e, tian2023resilient, chen2024heterogeneous, chang2022lamp}.
    Although they contributed to the provision of datasets underlying multi-robot research, they didn't offer a diverse set of benchmarks.
    In particular, they didn't include evaluation of place recognition, which is a key component of information exchange between robots. 
           
    \item \textbf{Multi-session} datasets were suggested the data acquired in day and night with the \ac{TIR} camera \cite{yun2022sthereo, lee2022vivid++}.
    Thus, they can report on the \ac{TIR} image different from the visible spectrum under low-light conditions. 
    However, they focused on structural environments.
    MCD \cite{nguyen2024mcd} is similar to our dataset, but their platform is limited to structured areas of the campus and does not provide \ac{TIR} images.

    \item \textbf{Multi-modal} datasets were offered with various sensor configuration \cite{jeong2024diter, geiger2012we, wang2024we, liu2024botanicgarden}.
    Wheeled robot dataset \cite{liu2024botanicgarden} provided its encoder data.
    Also, legged robot datasets including our previous work \cite{jeong2024diter, wang2024we} provided joint data from their platforms.
    Especially, TAIL \cite{wang2024we} has not been still expanded to a multi-robot dataset. 
\end{itemize}

\subsection{Ground Truth Generation}
Most datasets utilize a \ac{GPS} for providing the ground-truth \cite{jeong2024diter, geiger2012we, yun2022sthereo, wang2024we, huang2021disco, feng2022s3e}.
However, \ac{GPS} signals can be easily blocked in spaces such as underground or tunnels.
Also, it can be interfered with due to multiple reflections (i.e. multi-path) in built-up areas or regions surrounded by trees.
Especially, this phenomenon increases on campus, which coexists between a tower and a park.
Several datasets leveraged a coupling method between \ac{LiDAR} \ac{SLAM} and \ac{GPS} to generate more reliable ground-truth \cite{yun2022sthereo, lee2022vivid++, tian2023resilient}.
However, this approach struggles in narrow environments, such as corridors and alleyways.
To overcome these limitations, various approaches to generating ground-truth utilizing a survey-grade map were proposed \cite{chen2024heterogeneous, chang2022lamp, liu2024botanicgarden, nguyen2024mcd}. 

\citet{chang2022lamp} used the survey-grade map to generate the ground-truth utilizing a scan-to-map localization \cite{reinke2022locus}.
\citet{chen2024heterogeneous} employed a Leica RTC 360 laser scanner to build the indoor sequence of ground-truth.
It produces a highly accurate millimeter-level map, allowing them to generate precise ground-truth.
\citet{liu2024botanicgarden} also leveraged the RTC 360 and Leica Cyclone Register 360 software. 
Each scan is pre-registered by visual-inertial \ac{SLAM} and post-registered based on \ac{ICP} and pose graph optimization.
\citet{nguyen2024mcd} proposed continuous-time ground truth generation through survey-grade prior map registration, removing dynamic objects, sun glare, and reflection outliers to provide accurate ground-truth.

\begin{figure*}[t]
    \centering
    \includegraphics[clip, width=\textwidth]{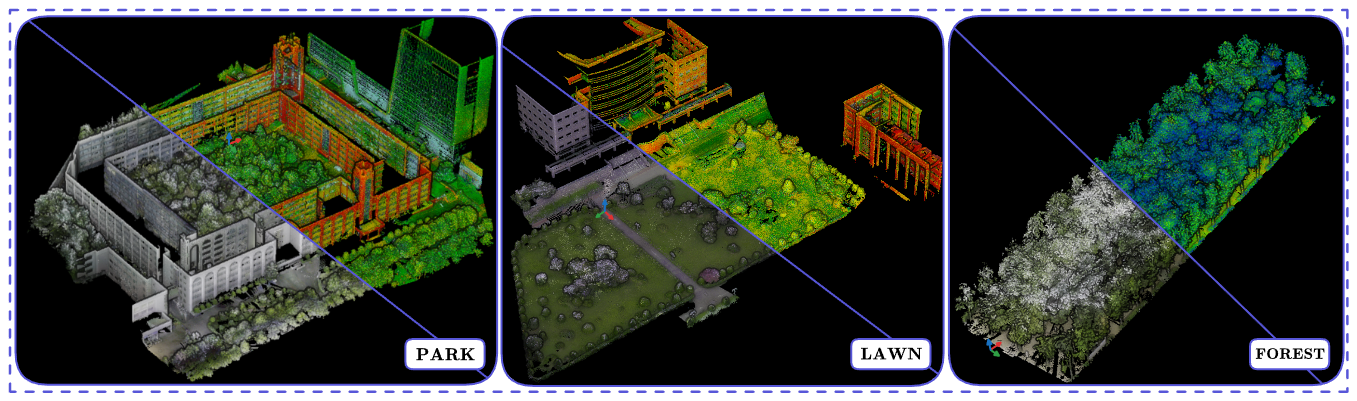}
    \caption{Obtained survey-grade map for each sequence. \texttt{PARK}, \texttt{LAWN}, \texttt{FOREST} maps are provided with accurate photo-metric representation, global intensity, and geometric scale obtained from survey device. For versatility in evaluation step, each survey-grade prior map are given its manual origin.}
    \label{fig:GT_maps}
\end{figure*}

\section{Sensor Configuration and Hardware}

\subsection{Sensor measurement, setup, and topics}
\begin{figure}[h]
    \centering
    \includegraphics[clip, trim= 0 120 0 130, width=0.48\textwidth]{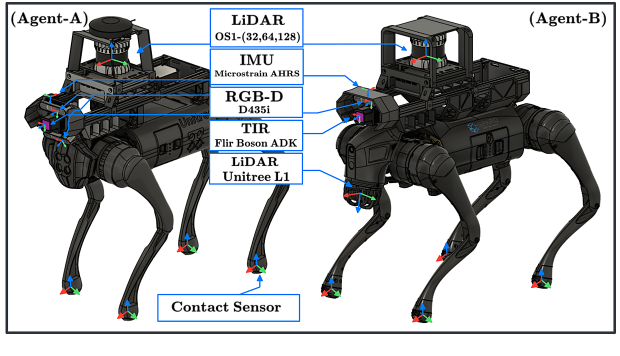}
    \vspace{-0.7cm}
    \caption{Illustration of robots with perceptual sensors and different ranging sensor setups. \textit{Agent-\textbf{A}} utilizes RGB-D for terrain information. \textit{Agent-\textbf{B}} utilized built-in \ac{LiDAR} sensor for geometric terrain information.}
    \label{fig:setup}
    \vspace{-0.3cm}
\end{figure}

As shown in \figref{fig:setup}, each robot shares a partially identical sensor setup, followed by different LiDAR (i.e. channel) and \ac{IMU}. 
Sensor specifications for each agent are configured in \tabref{tab:config}. 
Inheriting the main purpose of our previous work \cite{jeong2024diter}, both agents have the identical structure of a forward-looking sensor system consisting of RGB, and thermal cameras. 
Additionally, for a wide range of diversity in environment, measurement, and scenario, we also provide sequences with ground-facing LiDAR added agents.
The forward-facing sensors are positioned to simultaneously capture sufficient visual information as well as geometric and visual terrain data.

\begin{table}[h]
\caption{Sensor Specifications and Ros Topic Name}
\renewcommand{\arraystretch}{1.2}
\centering\resizebox{0.48\textwidth}{!}{\
\begin{tabular}{c|c|c|c|c}
\toprule
\hline
\textbf{Agent ID}        & \textbf{Hardware}            & \textbf{Sensors}          & \textbf{Specifications}                  & \textbf{Topic name} \\ \hline
\multirow{10}{*}{\textit{Agent-\textbf{A}}}& \multirow{6}{*}{Intel NUC} & \multirow{2}{*}{RGB-D}     & \multirow{2}{*}{Intel Realsense D435i}   & /agent0/ground/depth/image$\_$raw          \\ 
                         &                              &                           &                                          & /agent0/front/color/image$\_$raw         \\ \cline{3-5}
                         &                              & \multirow{2}{*}{Thermal}  & \multirow{2}{*}{FLIR Boson ADK}          & /agent0/flir$\_$boson/image$\_$raw         \\
                         &                              &                           &                                          & /agent0/flir$\_$boson/camera$\_$info       \\ \cline{3-5}
                         &                              & LiDAR                     & Ouster OS1-32                            & /agent0/ouster/points                          \\ \cline{3-5}
                         &                              & 9DoF-IMU                  & Microstrain 3DM-GX5-25                   & /agent0/imu/data                           \\ \cline{2-5}
                         & \multirow{3}{*}{Unitree-GO1} & Joint                     & \multirow{3}{*}{Built-in Robot}          &                                            \\
                         &                              &  6-DoF-IMU                &                                          & /agent0/high$\_$state                      \\
                         &                              &  Contact Sensor           &                                          &                         \\ \hline
\multirow{9}{*}{\textit{Agent-\textbf{B}}}& \multirow{5}{*}{Intel NUC} &{RGB}    & {Intel Realsense D435i}                  & /agent1/front/color/image$\_$raw         \\ \cline{3-5} 
                         &                              & \multirow{2}{*}{Thermal}  & \multirow{2}{*}{FLIR Boson ADK}          & /agent1/flir$\_$boson/image$\_$raw         \\
                         &                              &                           &                                          & /agent1/flir$\_$boson/camera$\_$info       \\ \cline{3-5}
                         &                              & LiDAR                     & Ouster OS1-64/128                            & /agent1/ouster/points                      \\ \cline{3-5}
                         &                              & 9DoF-IMU                  & Microstrain 3DM-GV7               & /agent1/imu/data                           \\ \cline{2-5}
                         & \multirow{4}{*}{Unitree-GO2} & Joint                     & \multirow{4}{*}{Built-in Robot}          &                                            \\
                         &                              &  6-DoF-IMU                &                                          & /agent1/high$\_$state                      \\
                         &                              &  Contact Sensor           &                                          & /agent1/go2$\_$lidar                        \\
                         &                              &  Built-in \ac{LiDAR}      &                                          &   \\
\bottomrule
\end{tabular}}
\label{tab:config}
\end{table}

Subsequently, the \ac{LiDAR} sensor is located at the rear side of the visual measurement mount, providing relatively accurate geometric measurement.
We conducted \textit{identical calibration procedure} from our previous work \cite{jeong2024diter} to each agent in order to acquire accurate extrinsic between sensors.   
Agent notation is done as shown in \tabref{tab:config} to distinguish acquired data with namespace, which can also be found in the obtained data topics. 

\section{DATASET}
\subsection{Sequence}
\begin{figure}[h]
    \centering
    \includegraphics[clip, trim=1200 320 1200 320, width=0.48\textwidth]{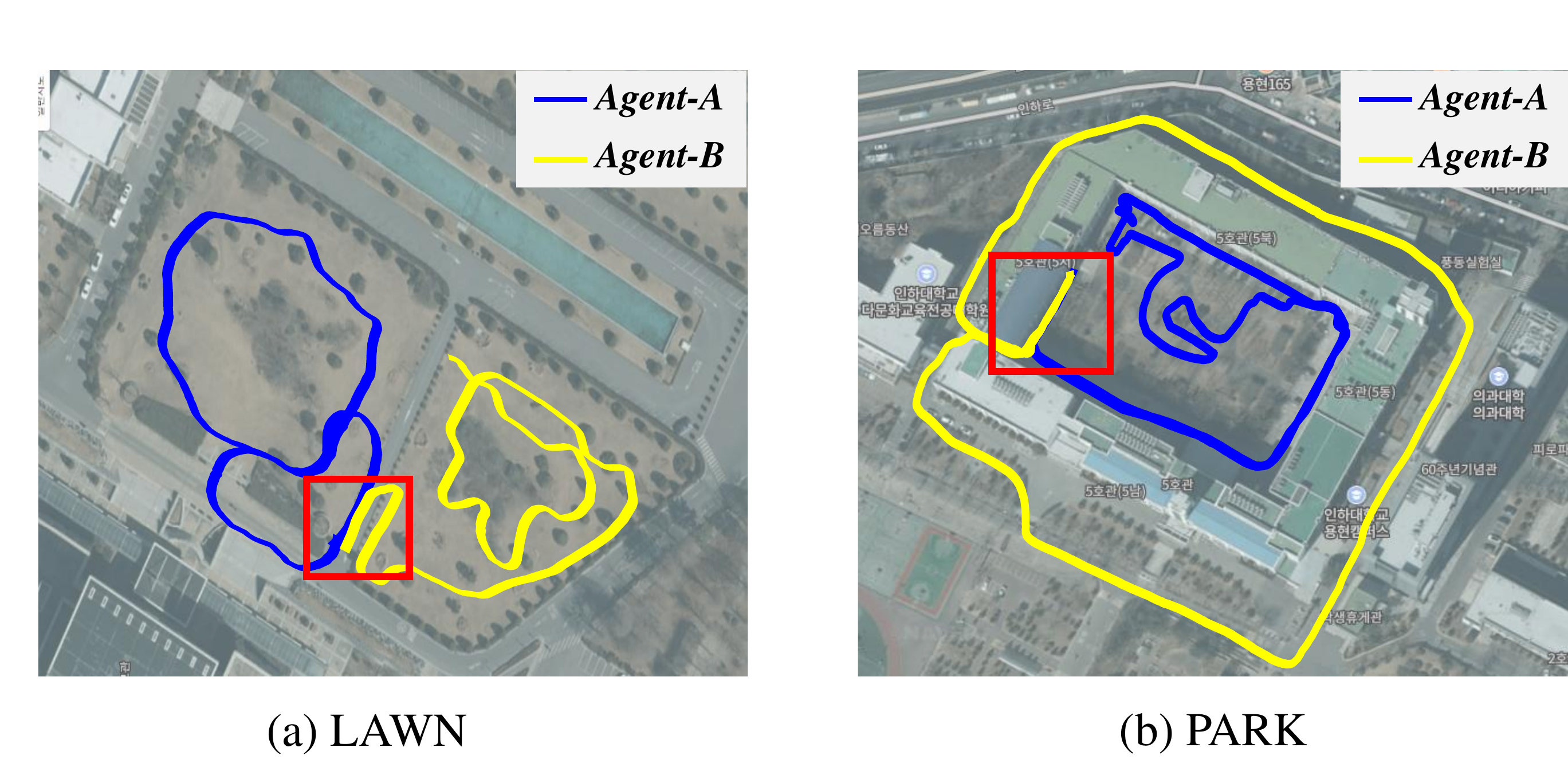}
    \caption{\texttt{LAWN} and \texttt{PARK} sequences are recorded by our multi-robot systems. \textit{Agent-\textbf{A}} (blue) and \textit{Agent-\textbf{B}} (yellow) traverse different regions for efficient mapping. Each sequence contains partial overlap in trajectory (red box).}
    \label{fig:gt_traj}
    \vspace{-0.4cm}
\end{figure}

\begin{figure*}[t]
    \centering
    \includegraphics[clip, trim= 70 30 50 0, width=\textwidth]{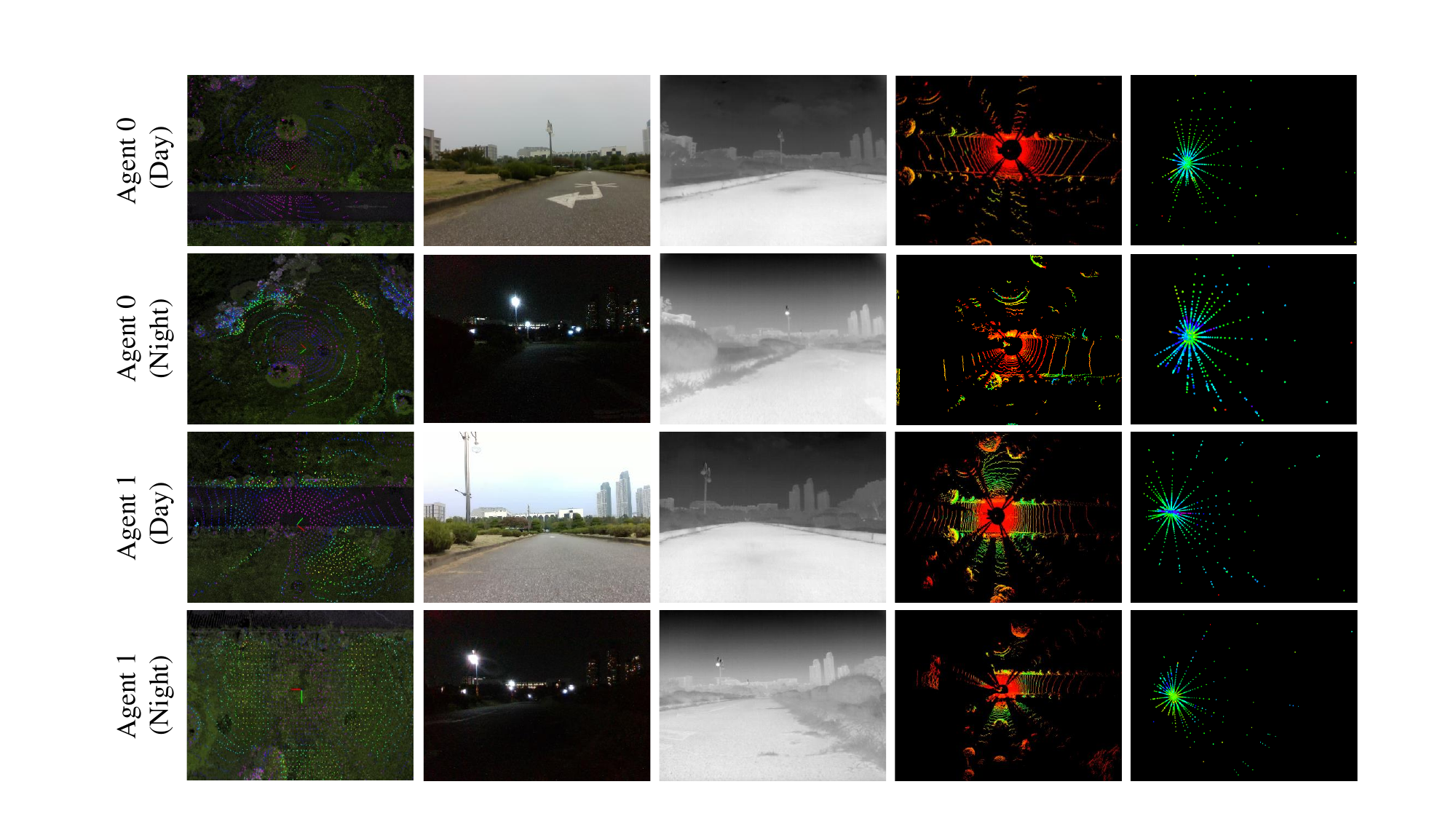}
    \caption{Data test on \texttt{LAWN} sequence with provided dataset and survey-grade prior-map. Starting from left, each row visualizes the point cloud using GT trajectory, RGB, \ac{TIR}, \ac{LiDAR}, and built-in \ac{LiDAR} from each agent, respectively.}
    \label{fig:data}
    \vspace{-0.4cm}
\end{figure*}

As represented in \figref{fig:gt_traj}, \figref{fig:forest_traj}, and \figref{fig:data}, all sequences are obtained within the outdoor sites and are conducted at different time and places, noted as \texttt{LAWN}, \texttt{PARK}, and \texttt{FOREST}, respectively. 
Each sequence includes multi-session and multi-robot scenarios, deploying agents with varying sensor configurations. We utilized two different platforms specifically noted as \textit{Agent-\textbf{A}} and \textit{Agent-\textbf{B}}, which vary in their sensor setup and locomotion characteristics. During data acquisition, we leveraged the advantages of legged-robots to traverse through such places with sudden height changes, where wheeled robots are unable to navigate.
\figref{fig:GT_maps}, which is a survey-grade map, visualizes the environment of our sequences in the abstract. The details of each sequence are as follows: 
 \begin{enumerate}
    \item \texttt{LAWN} is obtained from an extremely unstructured environment with terrains showing high levels of vegetation.
    \texttt{LAWN} sequence is covered by \textit{Agent-\textbf{B}} setup twice, which travels across the upper and lower region.
    Data collected from both agents capture significant motion and include scenes with occlusions caused by vegetation, as well as challenging traversal scenarios involving stones and narrow bushes.
    \item \texttt{PARK} is obtained around a campus building featuring a variety of environmental characteristics. 
    \texttt{PARK} sequence involves both \textit{Agent-\textbf{A}} and \textit{Agent-\textbf{B}} setup, covering areas from the inner region to the outer region. 
    The inner region of the \texttt{PARK} consists of a narrow underground area and unstructured terrain in the center, 
    while the outer region consists of areas with urban characteristics.
    PARK sequence includes scenes with multiple dynamic objects, indoor navigation, and continuous illumination changes. 
    \item \texttt{FOREST} inherits the same sequence from \cite{jeong2024diter}. However, with a gap of a year from our previous work,
    we deployed \textit{Agent-\textbf{B}} instead of an identical setup from our previous work to address a new challenge in both heterogeneous sensor application and long-term autonomy.
\end{enumerate}

\subsection{Survey-grade map}
In unstructured, narrow environments, generating ground-truth is a challenging task. By deploying legged-robot, our dataset contains multiple sequences that navigate GPS-denied areas such as indoors, under buildings, and in narrow structures. Additionally, multi-robot and multi-session tasks often require the evaluation of map results to assess the map's global scale and accuracy. We utilize survey-grade 3D LiDAR total station, RTC360 from leica to provide globally consistent ground-truth map and trajectory generation. The survey-grade map also provides spatial, photometric, and spectral information. To ensure the reliability of both the acquired survey-map and dataset, the survey-grade prior map for each site was acquired within 6 hours, and the dataset was acquired within 2 weeks after the survey-map generation.  

\subsection{Ground Truth Trajectory}
Previous works \cite{chen2024heterogeneous, chang2022lamp, liu2024botanicgarden, nguyen2024mcd} handle survey-grade data as ground-truth maps, and conducts scan-to-map matching of deskewed point using installed ranging and inertial sensors. 
Our dataset extends the following approach to multi-robot scenarios and extensively matches each point obtained simultaneously to the map. 
Unlike the wheeled robot and handheld mapping, legged-robots experience continuous dynamic motion during data acquisition, which hinders accurate motion estimation and LiDAR deskewing. 
This characteristic of the legged-robot can cause inaccuracies in trajectory estimation and scan-to-map registration, leading to the trajectory being unsuitable for use as ground truth.
Therefore, we utilized a modified version of Point-LIO \cite{he2023point} and Leg-KILO \cite{ou2024leg} and deskewed the obtained point to consider the dynamic motion of the legged-robot and give accurate motion-based localization on the survey-grade prior map. 
\begin{figure}[t]
    \centering
    \includegraphics[clip, trim= 0 110 0 110, width=0.48\textwidth]{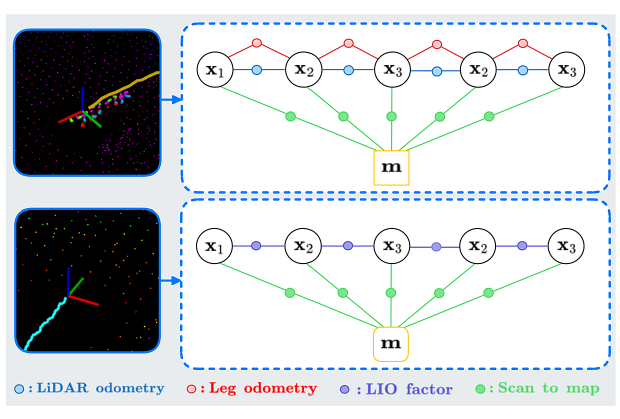}
    \vspace{-0.6cm}
    \caption{Tightly coupled factor-graph structure for our map-localization and ground-truth generation. Given factors are tightly coupled by scan-to-map registration factors in the optimization window.}
    \label{fig:gt_factor}
\end{figure}


\begin{figure}[t!]
    \centering
    \includegraphics[clip, trim= 0 142 0 140, width=0.48\textwidth]{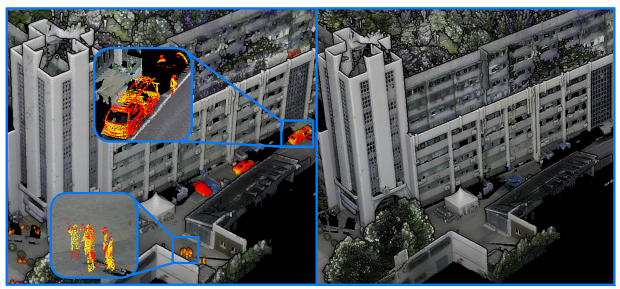}
    \caption{Original survey-grade map contains dynamic objects and outliers caused by pedestrians, vehicles, sun glare, and reflected points captured during the survey. To obtain an accurate trajectory, dynamic object and outlier removal were conducted. }
    \label{fig:dynamic_removal}
    \vspace{-0.4cm}
\end{figure}

However, challenges in generating ground truth trajectories still persist. Due to these difficulties in our dataset, localization on prior maps remains prone to drift and tracking failure. To address these issues and generate reliable ground truth, we employ a factor-graph-based localization framework based on GLIL \cite{koide2022globally, koide2024tightly} as shown in \figref{fig:gt_factor}. The following localization framework shows robust localization capability even in sudden changes in space, ensuring the reliability of ground-truth trajectory. Furthermore, localization on the prior map can still be hindered by dynamic objects and unreliable points caused by sunlight and reflection from glass and water. As shown in \figref{fig:dynamic_removal}, these elements are thoroughly removed by handcrafted segmentation to maximize scan-to-map matching during ground-truth trajectory generation.

\subsection{Challenges in DiTer++}
\textbf{Dynamic Motion:} The challenging aspect of the legged-robot platform arises from dynamic motion during its locomotion.
\begin{figure}[H]
    \centering
    \includegraphics[clip, trim= 0 15 0 15, width=0.48\textwidth]{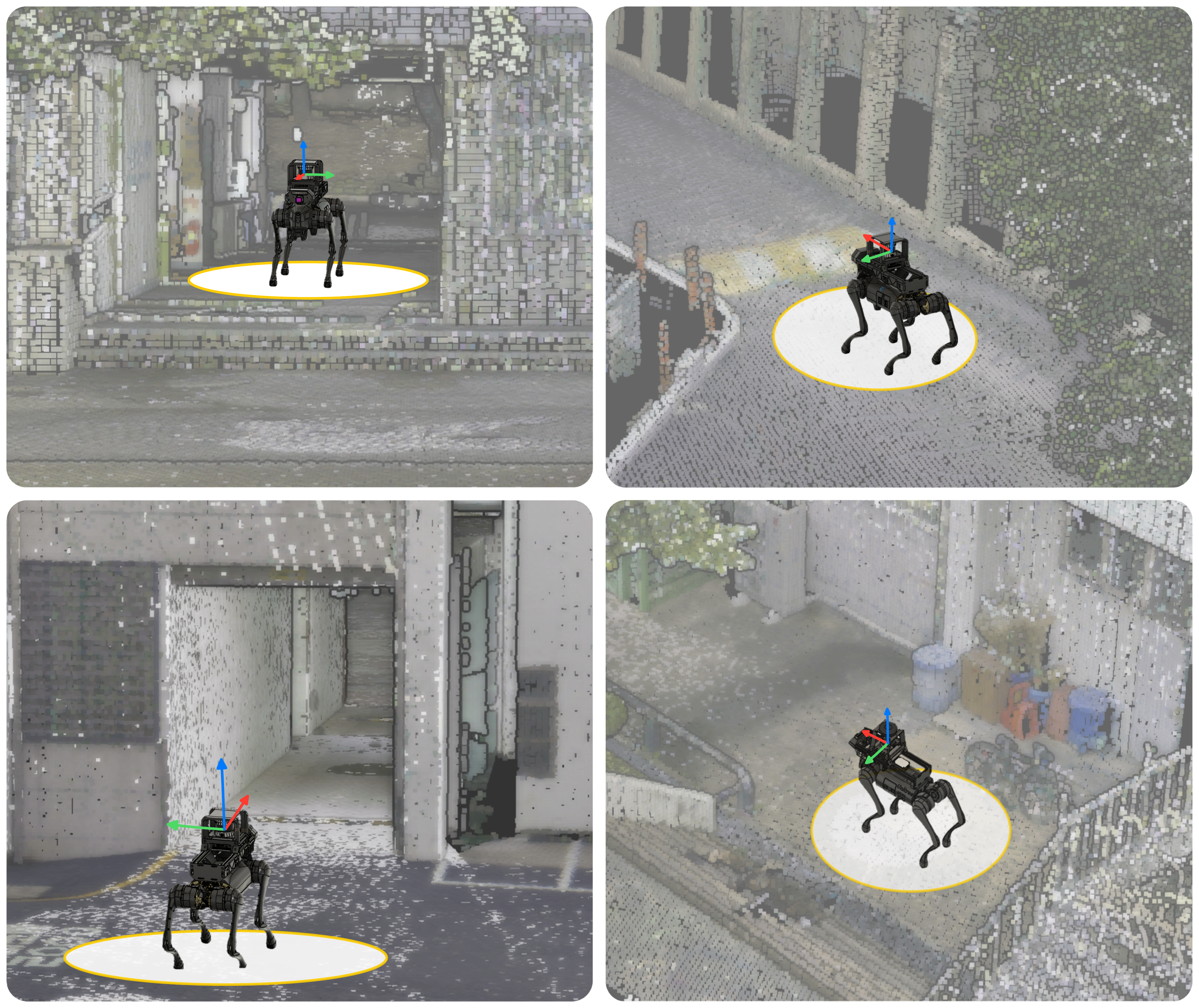}
    \caption{Example of narrow corridors that can be found in \texttt{LAWN} sequence. The legged-robot navigates through these narrow corridors by stepping over stairs, bumps, and paved terrain.}
    \label{fig:narrow}
    \vspace{-0.2cm}
\end{figure}
Dynamic motions during locomotion cause incremental drift in inertial measurement over time. 
Long-term data acquisition of our dataset contains this aspect of continuous stimulus to an inertial measurement against every DoF, especially against the Z-axis. Additional challenges from dynamic motion are likely to be found in exteroceptive observations. Tracking failure in visual measurement or noisy deskewed point clouds from dynamic motions are also inevitable, which can lead to inaccurate estimation such as pose estimation or accurate mapping result. Since our dataset navigates through various conditions by fully exploiting the legged-robot's traversability, our dataset is likely expected to confront various challenges that can be addressed in legged-robot autonomy in the fields.

\begin{figure*}[t]
    \centering
    \includegraphics[clip, trim= 0 0 0 0, width=\textwidth]{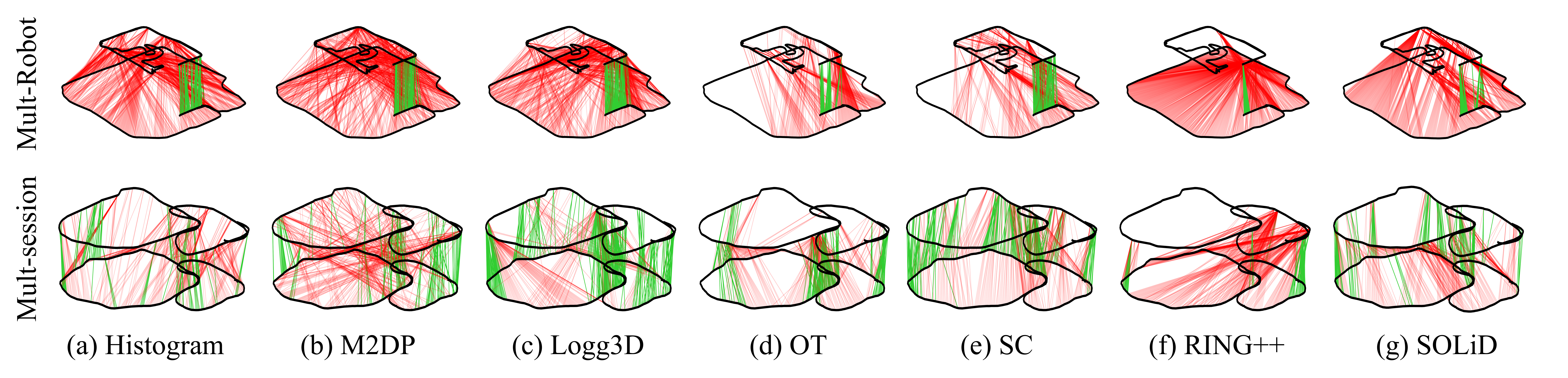}
    \caption{Top is multi-robot place recognition's 3d graph in the \texttt{PARK}. Bottom is multi-session place recognition's 3d graph in the \texttt{LAWN}. Red lines represent false matching loops and green lines represent true matching loops. The baseline for the true matching loops is when the two matched positions are within 10 meters.}
    \label{fig:3d_graph}
    \vspace{-0.2cm}
\end{figure*}
\begin{figure*}[h]
    \centering
    \includegraphics[clip, trim= 0 40 0 0, width=1.0\textwidth]{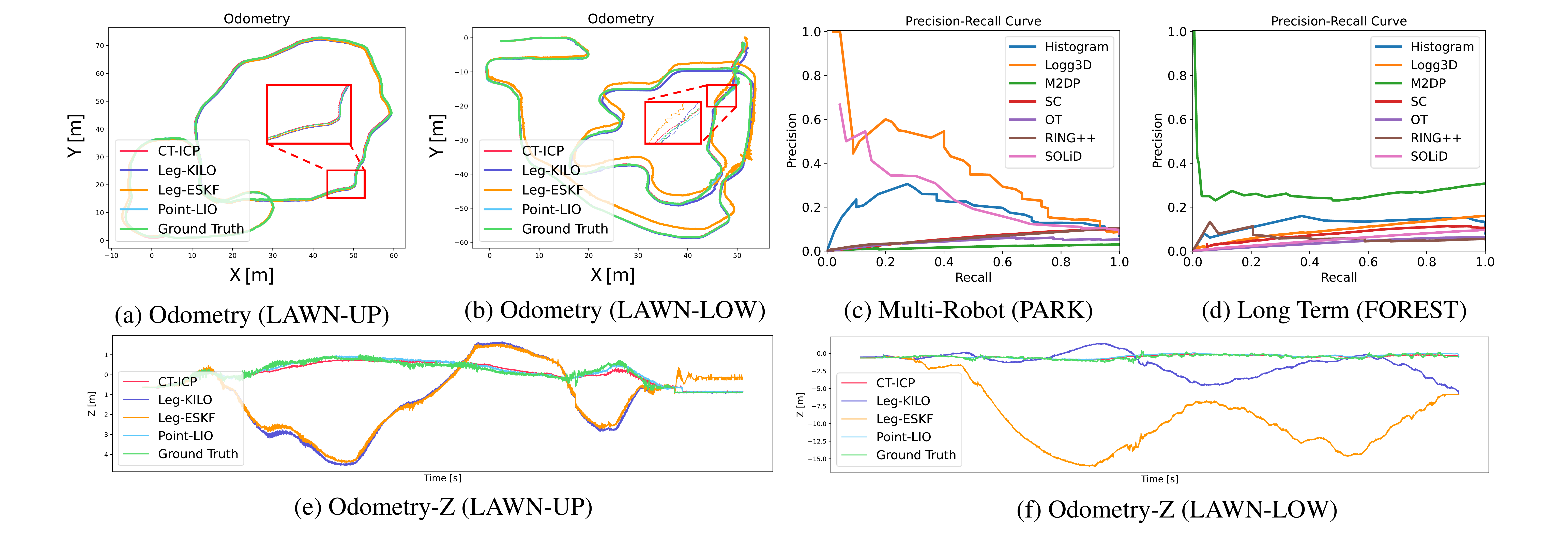}
    \caption{Odometry Evaluation. (a) Odometry results in \texttt{LAWN}'s \texttt{upper} sequence. (b) Odometry results in \texttt{LAWN}'s \texttt{lower} sequence (c) PR curve in the \texttt{PARK}'s \texttt{Night} sequence. (d) PR curve in the DiTer and DiTer++'s \texttt{FOREST} sequence. Challenging characteristics from sequences are reflected in odometry. Compared to \ac{LiDAR}-only methods, Inertial and kinematic-based odometry shows an extreme change in elevation and jitters in overall trajectory results as (e) and (f).}
    \label{fig:evaluation}
    \vspace{-0.2cm}
\end{figure*}

\begin{figure}[h!]
    \centering
    \includegraphics[clip, trim= 0 0 0 0, width=0.48\textwidth]{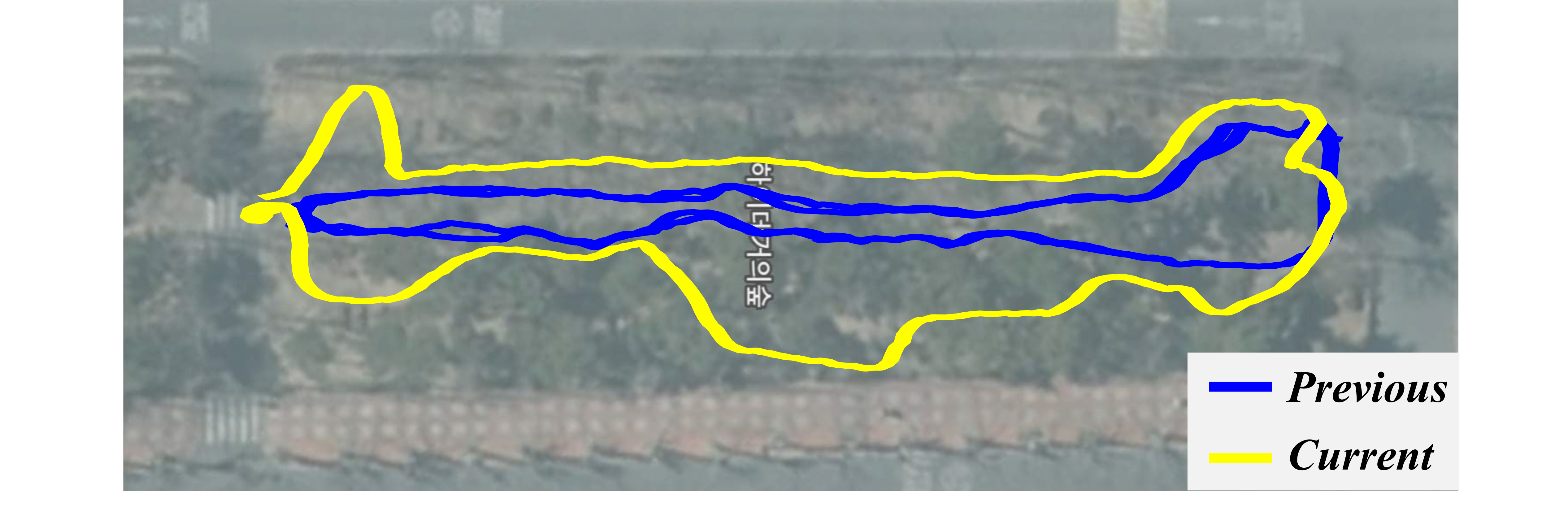}
    \caption{ Blue trajectory is the ground-truth of the DiTer's \texttt{FOREST} sequence obtained in \texttt{2023-05-01}. Yellow trajectory is the ground-truth of the DiTer++'s \texttt{FOREST} sequence obtained in \texttt{2024-09-11}. }
    \label{fig:forest_traj}
    \vspace{-0.6cm}
\end{figure}

\textbf{Narrow Scene:} Our dataset contains various terrains and challenging scenes evenly distributed over each and every site.
To be noted, the \texttt{PARK} sequence contains various narrow scenes (e.g. alleyway, corridoor) as shown in \figref{fig:narrow}.
These scenes contain high chances of failure in \ac{LiDAR}-based tasks such as odometry or \ac{SLAM}.

\subsection{Tandem with DiTer}
In the \texttt{Forest} sequence, we maintain a tandem with our previous work, DiTer \cite{jeong2024diter} as shown in \figref{fig:forest_traj}.
Even after about a year, we can extract pose ground-truth from the survey-grade prior map and robust scan-to-map-matching.
Since each robot has opposite travel directions and was acquired in different seasons, two sequences can be utilized to study challenging tasks (e.g. long-term place recognition).

\section{Benchmarks}
We evaluated our datasets in two main categories (i) odometry and (ii) place recognition.
Because our dataset's vision sensor is a monocular setup, the odometry derived from RGB and \ac{TIR} cameras suffer from a scale ambiguity.
To provide the metric-level benchmarks, we performed evaluations focusing on \ac{LiDAR}.

\subsection{Odometry}
To perform metric-level benchmarks on our dataset, we selected the following algorithms \cite{dellenbach2022ct, ou2024leg, he2023point} to verify one of our dataset's research versatility. Ranging, inertial, kinematic sensors are utilized in our odometry evaluation and are thoroughly compared against our ground truth trajectory as shown in \figref{fig:evaluation}.



\subsection{Place Recognition}
Each robot needs to recognize the other's revisited place to exchange information, so place recognition plays an essential role in multi-robot \ac{SLAM}.
Therefore, we provide various LiDAR-based place recognition approaches \cite{rohling2015fast, he2016m2dp, kim2018scan, vidanapathirana2022logg3d, ma2022overlaptransformer, xu2023ring++, kim2024narrowing} as benchmarks.
\figref{fig:3d_graph} presents a 3d graph for checking the multi-robot and multi-session place recognition performance.
All of the methods used in benchmarks suffer from recognizing revisited places.
Logg3D, SOLiD, and Histogram perform well in multi-robot PR curves as shown in \figref{fig:evaluation}.
On the other hand, long-term place recognition shows a decrease in the performance of SOLiD and Histogram based on the number of points.

\section{CONCLUSION}
We propose a diverse terrain, multi-modal, multi-robot, and multi-session dataset called DiTer++, which is the evolved version of our previous work \cite{jeong2024diter} utilizing legged-robot. In our future work DiTer$\#$, our dataset will contain actual multi-robot communication scenarios, additional platforms majorly consisted of legged-robots, and campus-scale data acquisition followed with identical survey-grade map acquired at a full campus size. 

\end{document}